\documentclass[letterpaper, 10 pt, conference]{ieeeconf}  

\IEEEoverridecommandlockouts                              

\usepackage{graphics} 
\usepackage{epsfig} 
\usepackage{mathptmx} 
\usepackage{times} 
\usepackage{amsmath} 
\usepackage{amssymb}  
\usepackage{graphicx} 
\usepackage{graphicx,subfigure}
\usepackage{algorithm}
\usepackage{algpseudocode}
\usepackage{threeparttable}

\title{\LARGE \bf  Learning needle insertion from sample task executions}

\author{Amir Ghalamzan E.$^{1\dagger}$
\thanks{$^\dagger$ $^1$University of Lincoln, Intelligent Manipulation Lab (IML)
       \texttt{aghalamzanesfahani@lincoln.ac.uk} }
}

\begin{document}

\maketitle
\thispagestyle{empty}
\pagestyle{empty}

\begin{abstract}
Automating a robotic task, e.g., robotic suturing can be very complex and time-consuming. Learning a task model to autonomously perform the task is invaluable making the technology, robotic surgery, accessible for a wider community. The data of robotic surgery can be easily logged where the collected data can be used to learn task models. This will result in reduced time and cost of robotic surgery in which a surgeon can supervise the robot operation or give high-level commands instead of low-level control of the tools. We present a data-set of needle insertion in soft tissue with two arms where Arm 1 inserts the needle into the tissue and Arm 2 actively manipulate the soft tissue to ensure the desired and actual exit points are the same. This is important in real-surgery because suturing without active manipulation of tissue may yield failure of the suturing as the stitch may not grip enough tissue to resist the force applied for the suturing.
We present a needle insertion dataset including 60 successful trials recorded by 3 pair of stereo cameras. Moreover, we present Deep-robot Learning from Demonstrations which predicts the desired state of the robot at $t+1$ (which is due to the optimal action taken at $x_t$) by looking at the video of the past time steps, i.e. $t-N$ to $t$ where $N$ is the memory time window, of the task execution. 
The experimental results illustrate our proposed deep model architecture is outperforming the existing methods. 
Although the solution is not yet ready to be deployed on a real robot, the results indicate the possibility of future development for real robot deployment. 

\end{abstract}

\section{Introduction}
Robotic suturing is an important task during robotic surgery which may require significant time. Fully tele-operating such tasks may be tedious and impose a high cognitive loading on the surgeons. 
Robot learning from demonstrations (LfD)~\cite{ravichandar2020recent} can be exploited to learn a task model to automate such tasks. 
Automating tasks, e.g. in robotic surgery, can significantly reduce fatigue and cognitive load on the surgeon resulting in improved performance~\cite{moller2020laparoscopic}. 
Conventional LfD approaches are only useful to capture some common patterns in the observed trajectories where they can be generalised to different environment configurations, a novel goal point, such as DMP, or several via points, such as ProMP. In such cases, high-level feature are hand coded.
Recent developments in deep learning allows feature extraction using only a dataset of the sample task. 
Hence, deep LfD (D-LfD) can be developed to learn much more complex tasks which requires highly complex feature extraction from images of the scene.

\begin{figure}[tb!]
\subfigure[]{\label{fig:schNI1}\includegraphics[width=.25\columnwidth]{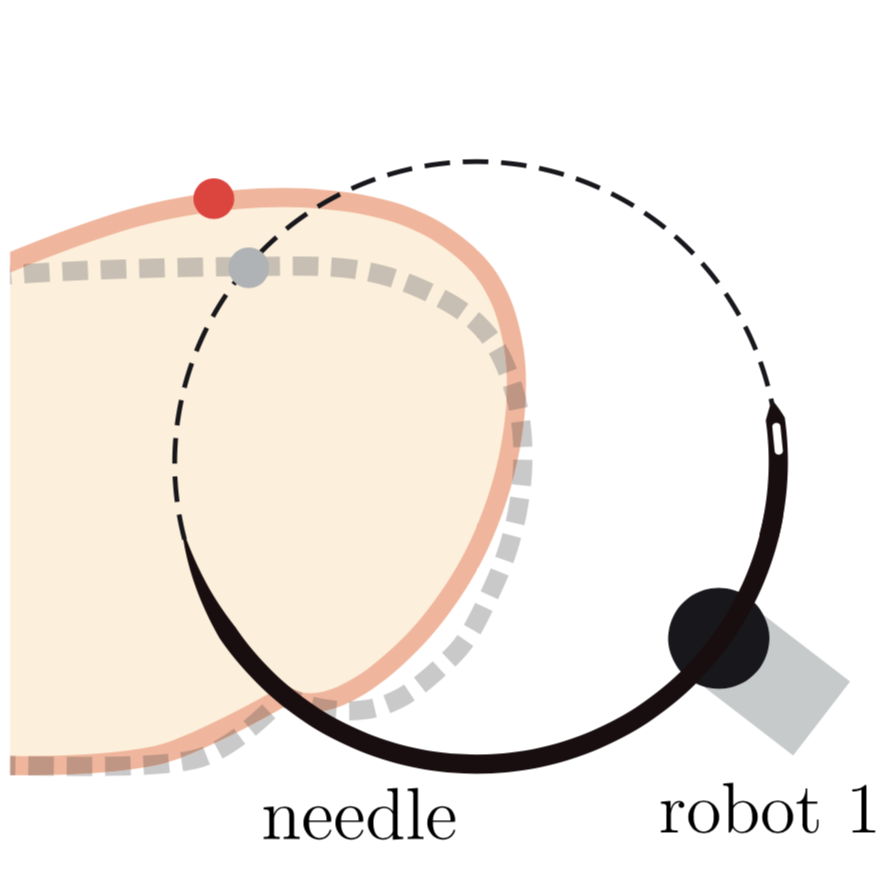}}
\subfigure[] {\label{fig:schNI2}\includegraphics[width=.25\columnwidth]{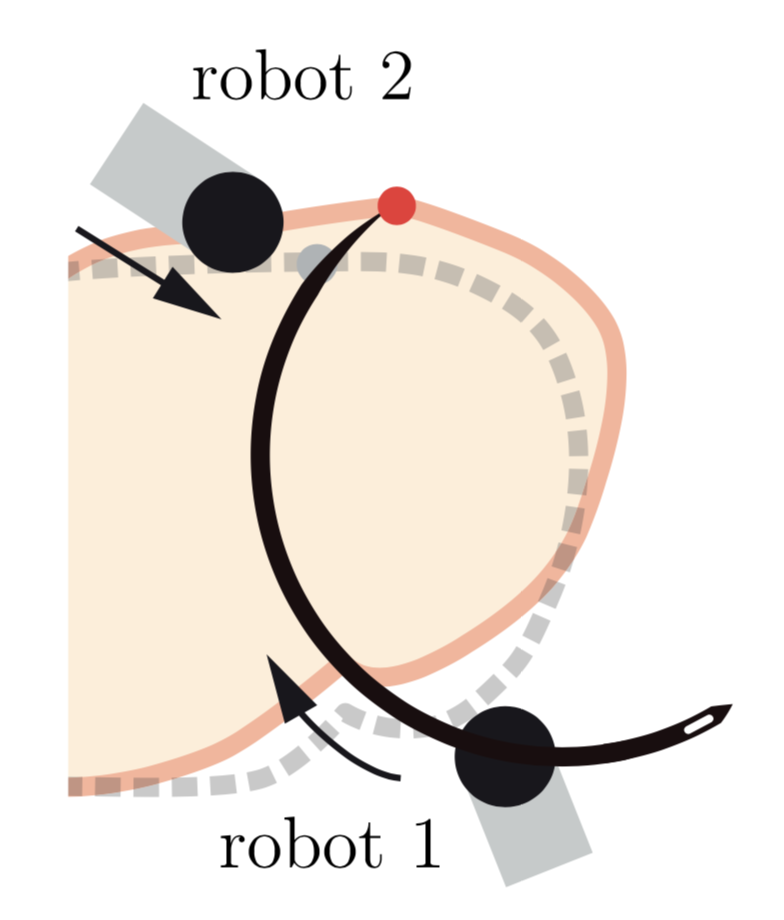}}
\subfigure[]{\label{fig:insertion}\includegraphics[width=.4\columnwidth]{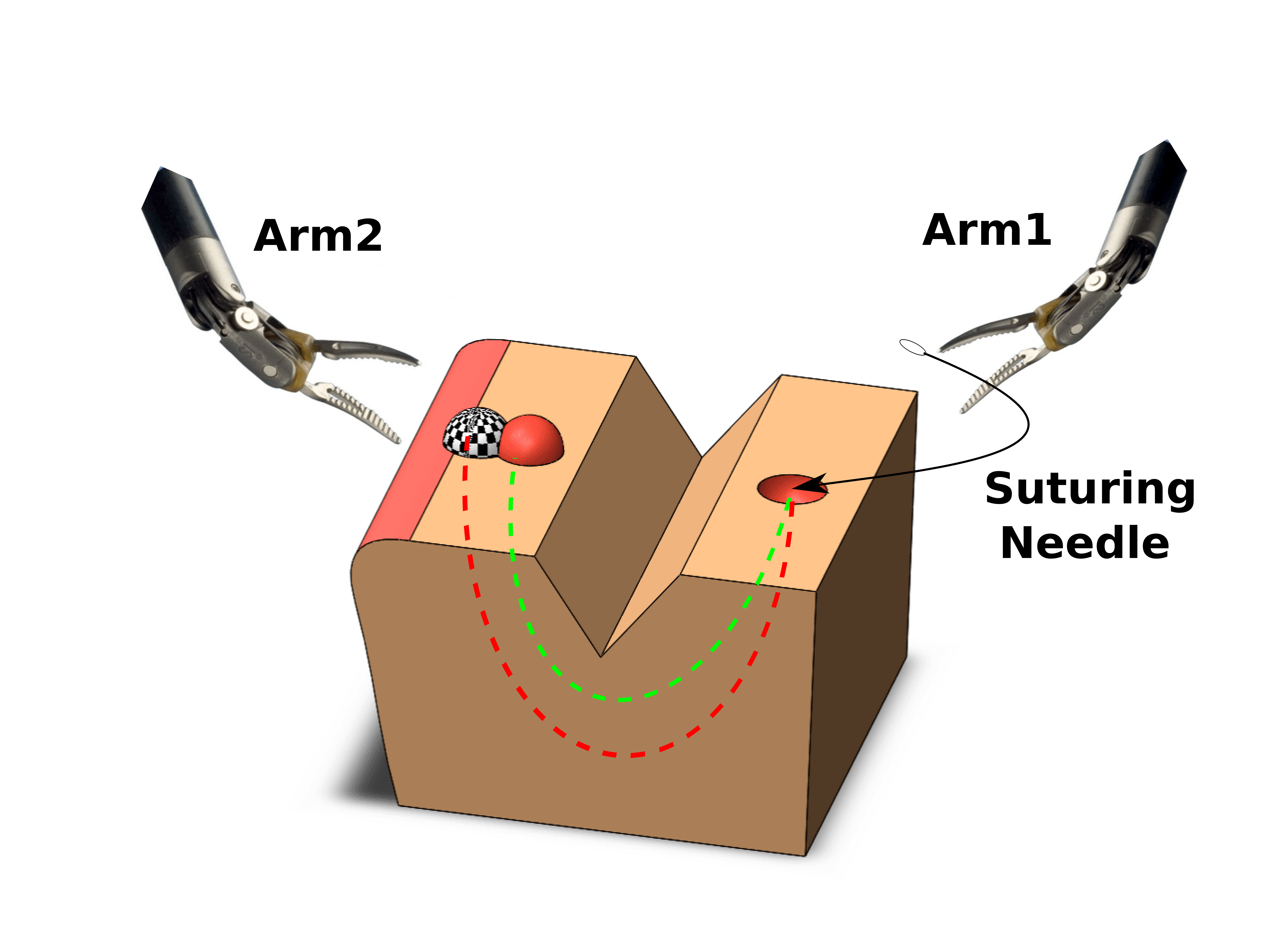}}\\
\centering
\subfigure[]{\label{fig:setup}\includegraphics[width=.47\columnwidth]{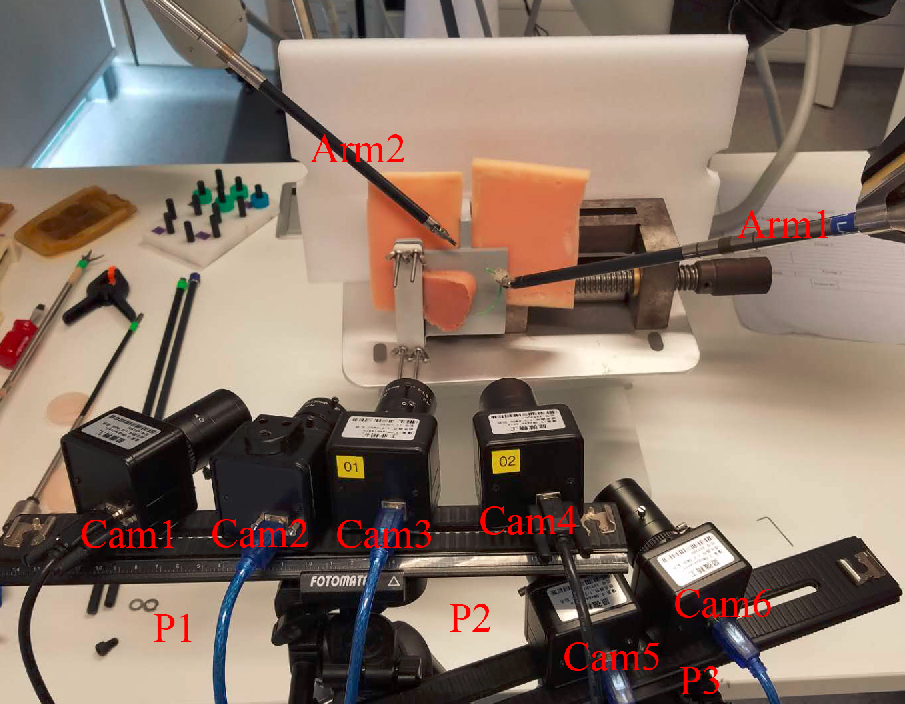}}
\caption{Needle insertion in soft tissue: a 2D schematic of tissue deformation due to needle insertion--original shape shown as grey dash line in \subref{fig:schNI1}; hence, the target exit point (red dot in \subref{fig:schNI1}/black-white sphere in \subref{fig:insertion}) differs from the actual one (grey dot in \subref{fig:schNI1}/red sphere in \subref{fig:insertion}); 
the needle pushes the tissue in \subref{fig:insertion} making the actual (red sphere) and desired (black-white sphere) needle exit points different--the actual (green dashed line) and desired (red dashed line) paths of the needle;
Arm 2 actively manipulates the tissue in \subref{fig:schNI2} to ensure the actual and target exit point are the same; dataset collection setup (\subref{fig:setup}).  
}
\label{fig:insexp}
\vspace{-.5cm}
\end{figure}


%
%

Continuous Hidden Markov Models~\cite{power2015cooperative}, Dynamic Movements Primitives~\cite{yang2018learning} and Model Predictive Control~\cite{shin2019autonomous} are used to automate needle passing trajectory, suturing trajectories and tissue manipulation. However, hand designing control model or features are time-consuming, and not seamlessly generalisable. 
Learning a task model from some sample task executions is proposed to address this issue, e.g. imitation learning, Learning from Demonstrations, Apprenticeship Learning and Inverse Optimal control that may require hand designed features limiting their generalisability. 
%

%
For example, Sharma \textit{et al.} \cite{sharma2019third} developed a third person imitation learning algorithm which has a separated goal detection and low level control part. The goal detection part makes the overall control loop task-dependent and cannot readily be generalised to other tasks. 
Watter~\textit{et al.}~\cite{watter2015embed} defined constraints on a variational auto-encoder (VAE) to guarantee a locally linearisable latent space suitable for implementing optimal control policy. 
However, we  nee a longer term dependencies for a reliable controller. 

Ensemble learning is a process of effectively combining multiple models in such a way to improve the overall performance of a predictor \cite{webb2004multistrategy} and have better evaluation for prediction on unseen data \cite{muhlbaier2005ensemble}. \cite{dantas2018improving} used bagging with exponential smoothing for improving time series forecasting. \cite{chen2009bagging} implemented bagging to obtain better robustness and accuracy with Gaussian process regression models. They used weighted averaging weighted by inverse of the predictive variance. \cite{chua2018deep, kurutach2018model} enhanced the robustness of deep model-based reinforcement learning by bagging ensembles.

Becker \textit{et al.}~\cite{becker2019recurrent} proposed a Kalman filter network (KFN) in the latent space which takes auto-encoded high dimensional input data whereas the filtering process is simplified by avoiding matrix inversion calculation by assuming diagonal covariance matrices. In contrast to KFN, D-LfD benefits from CNN, inherently more powerful feature extractor without any simplification assumption whereas the robot kinematic and calibration data are also fed to the network in latent space.  
Probabilistic methods outperform deterministic approaches in dealing with modelling errors, non-modelled disturbances, and deficient sensor data in various applications. EKFs are efficiently used to train a NN where it does not suffer from the problem of being stuck in poor local optima~\cite{siswantoro2016linear,de2017stable} as it uses second order derivatives in weight update step. 
%
%
The proposed networks are not demonstrated in real-world experiments due to limited accurate because of the fixed linear transition model in calculations of the filtering has simple architecture. %
In contrast, our novel controller design, i.e. D-LfD, uses recurrent network after feature extraction and shows a good accuracy to predict the control input for a highly complex and nonlinear task of needle insertion in a deformable soft object.

The main contribution of this paper is a control architecture for Deep Imitation Learning that learns the control policy of a complex task of inserting a needle into a deformable object by da Vinci, a surgical robot, from a data-set of demonstrations. 
We are proposing a novel control system design for a robot: 
we present a needle insertion data-set-- a da Vinci robot inserting a semi-circular needle into a deformable object by Arm 1 while Arm 2 is actively manipulating the object to ensure the desired and actual exit points are the same. 
Moreover, we propose a baseline D-LfD (Recurrent Multi Layer Perceptron (RMLP) trained by a (Extended) Kalman filter--KF-RMLP)-- see Fig.~\ref{Fig::din}--that learns to generate the control actions for $t$ to reach the desired state of the system at $t+1$ given the the robot states and images of task execution in a time window of $t-N$ to $t-1$. 
%
Our experiments shows KF-RLMP outperforms the Feed-forward, RNN, GRU and LSTM in terms of precision of action prediction for the needle insertion task. 
KF-RMLP, also generates confidence bounds for the predicted actions where the confidence bounds can be used to generate a proper haptic-guidance to ease the complex teleoperation tasks for the operator. 
For high consequence and safety critical tasks, e.g. robotic surgery and self-driving cars, it is desired to keep the human operator in charge of the task execution whereas teleoperation can be eased either via shared control of the task~\cite{abi2019haptic} or giving haptic guidance indicating optimal actions to the operator~\cite{selvaggio2019haptic}. 

\section{Problem Formulation}

\subsection{Needle insertion task}
In this paper, our experiments include manipulation of deformable objects for needle insertion/suturing with two arms of a minimally invasive robot (shown in Fig.~\ref{fig:insexp}). 
Inserting the circular needle in (and piercing it thorough) a soft tissue deforms the tissue such that not enough tissue supports the stitch which may result in failure as the stitch may cut the tissue. 
In robotic surgery, Arm 2 (Fig.~\ref{fig:schNI2} and \ref{fig:insertion}) is used to manipulate the soft tissue to ensure the needle passes/exits through the desired point yielding enough gripped tissue by the stitch. 
In the common/current practice, a surgeon teleoperates (i) Arm 1 of the robot in a remote workspace (see Fig.~\ref{fig:schNI1}) to grab the needle and insert it into the object and (ii) Arm 2 to manipulate the deformable tissue (see Fig.~\ref{fig:schNI2}) to ensure needle passes through the desired exit point (the red point on Fig.~\ref{fig:insertion}). 
This procedure is delicate and requires significant training by novice surgeons and automating this by Deep-LfD is of notable interest. 
We have created a dataset ({\tt {https://github.com/imanlab/d-lfd} }) of inserting a needle into a deformable object using a da Vinci Research Kit (DVRK) shown in Fig.~\ref{fig:insexp}.

\subsection{Deep Learning from Demonstrations (D-LfD)}

Consider an intelligent agent, $\mathcal{M}$, observing the state of a system via a few cameras fully expressing the interaction of a robot with its environment in a remote workspace. 
We also assume the agent has a good knowledge about the kinematic of the robot as well as the relative pose of the cameras w.r.t. the robot base frame.  
We are interested in the intelligent agent that generates control action at each time $a_{t}$ for the robot to successfully complete the task. 
At each sampling time step $t$, the dataset includes $\theta_{l,t}\:\&\: \theta_{r,t} \in \mathbb{R}^{(1{\times}6)}$: kinematic data of Arm 1 and Arm 2 in joint space; $ee_{l,t}$ and $ee_{r,t}$: pose of Arm 1 and Arm 2 in their respective base frame where $ee\in SE(3)$\footnote{where $SE(3)$ denotes  the  group of 3D poses including 3D position $p$ and 3D orientation $q$; $SE(3)$  $=$ $R^3\times SO(3)$, $p\in R^3$ and $q\in SO(3)\subset R^{3\times3}$}; $a_{l,t}$ and $a_{r,t}$: control inputs for Arm 1 and Arm 2 at time $t$-- here, we have a low level PID controller which ensures the robot reaches the $\theta_{t+1}$, hence, $a_{t} = (ee_{t+1} - ee_{t})\in SE(3)$; $g\in \mathbb{R}^{2\times6}$: 2D tracking target point on the image captured by 6 cameras; $h\in \mathbb{R}^3$: computed 3D position of the target point w.r.t Camera 3.

Videos of all task demonstrations are captured by different cameras (as shown in Fig.~\ref{fig::videtimelapse}). 
We express the robot's states by ${s}_t= \{ee_{l, t},\: ee_{r, t}\}$ at sampling time $t$. 
Hence, our observation is videos of task execution $\mathrm{O}$ which are sampled to generate a sequence of images of the task according to the control frequency, i.e., $\mathbf{O}_i =\{o_{i,0}, o_{i,1},. . ., o_{i,T}\}$ where $i$ denotes the camera number, $o_{i,t}$ is the image at time $t$ corresponding to $\{{s}_t, {a}_t\}$, 
$\mathbf{s}=\{{s}_0,. . .,s_T\}$, $\mathbf{u} = \{a_t, . . ., a_{T-1}\}$ and $T$ is time-to-completion. 
We also assume the calibration parameters of each camera w.r.t. the base frame of Arm 1, denoted by $\mathbf{T}^c_i$, is also available.
A dataset of task (denoted by ${D} = \{\zeta_1, . . .,\zeta_{N_{demo}}\}$ and demonstrated by experts) is available where each demonstration, $\zeta_d = \left\{\{\mathbf{O}, \mathbf{s}, \mathbf{u} \}, \mathbf{T}^c\right\}_d\:\forall \: d=1, . . .,N_{demo}$, is the optimal solution to an unknown reward function corresponding with environment $E_d$, where $N_{demo}$ is the number of demonstrations.

We are interested in a model learning the mapping between the observations and actions from the corresponding successful task demonstrated data, $D$. 
Our control problem can be defined by a state space $S \subset \mathbb{R}^n$, an action space ${A} \subset \mathbb{R}^m$, a state transition function $T (s \in S , a \in {A}): \mathbb{R}^{n + m} \rightarrow \mathbb{R}^n$.
In the next section, we present Deep Learning from Demonstrations (D-LfD) model that computes the optimal action $a_t \in {A}$ based on the robot state $s_t \in {S}$ and the camera observations $o_{i, t}$, for a new unseen task execution/environment in the training set.

\begin{figure}
  \begin{center}
    \includegraphics[width=0.48\textwidth]{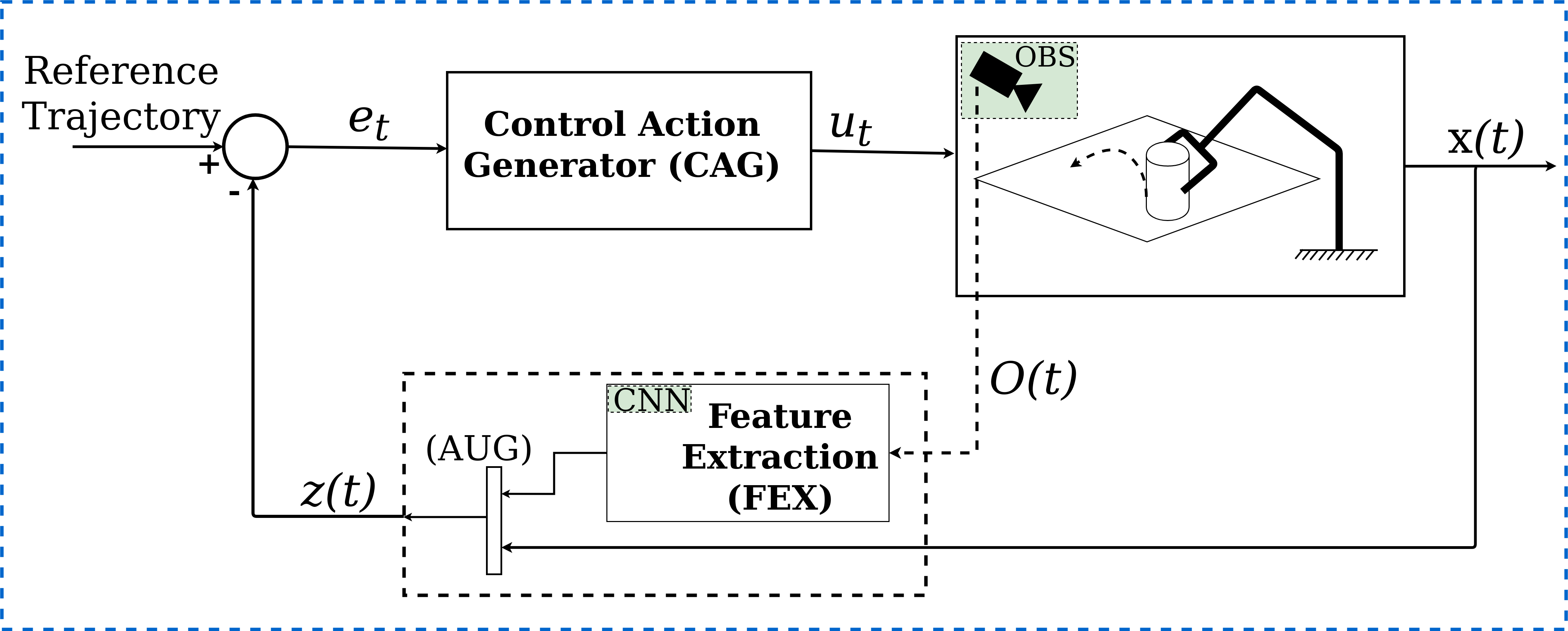}
  \end{center}
  \caption{A schematic of conventional controller with hand-designed features and controller structure.}
  \label{fig::concontroller}
\end{figure}

Conventional closed-loop control systems (Fig.~\ref{fig::concontroller}) use a state estimator generating desired states, e.g. the object position, from camera observations and a controller which generates robot's actions based on the desired target pose. 
While discriminative models can be used to estimate the system's state from high dimensional video observation, e.g. tracking an object in robot's workspace, they are built based on hand-designed state space, hence, they are limited in terms of generalisations. 

\begin{figure}
\vspace{-.5cm}
\includegraphics[trim=0cm 0cm 0cm 9cm, clip=true, width=1.03\linewidth]{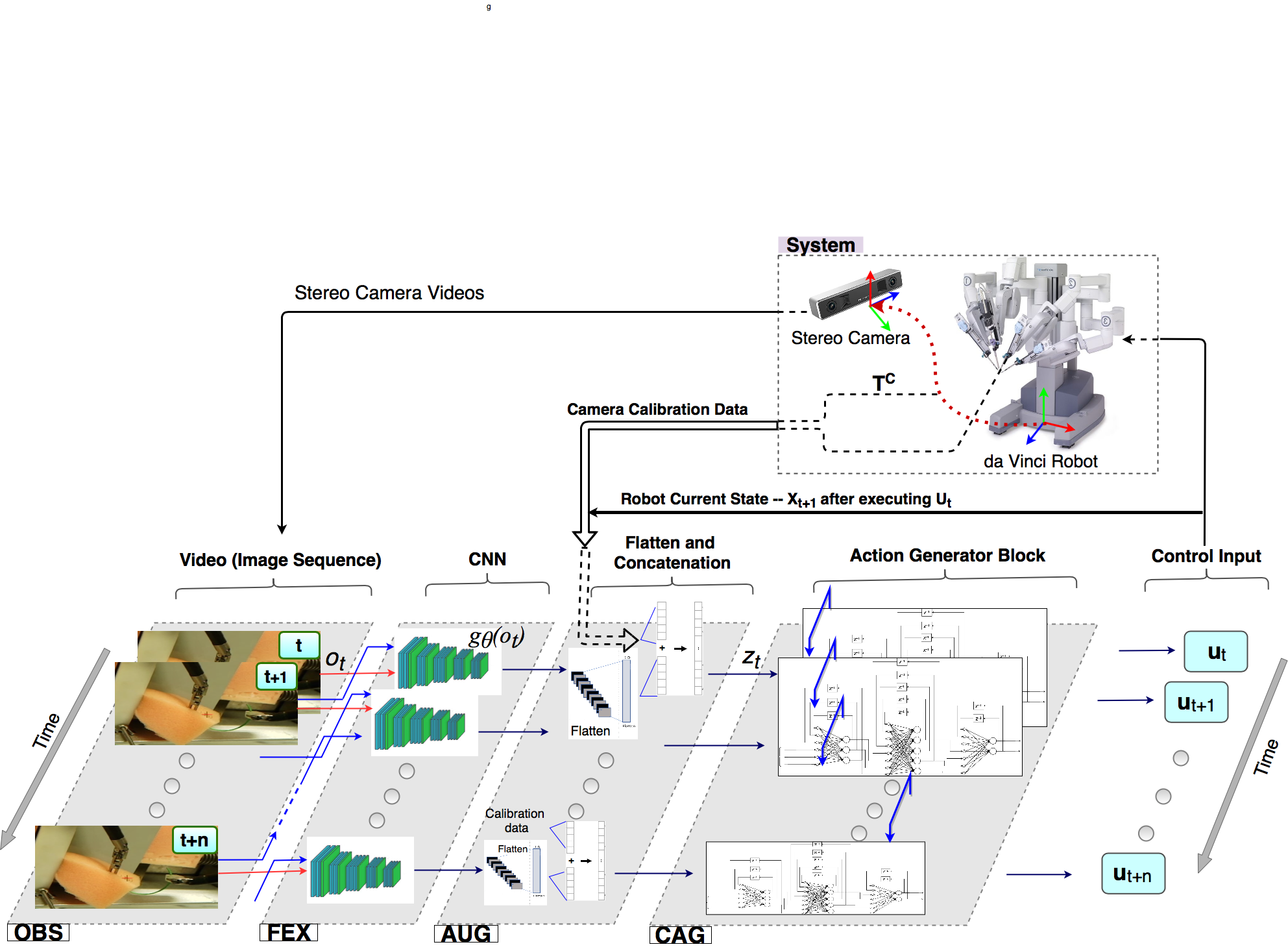}
\caption{Illustration of network architecture. The whole observations $\textbf{O}$ are propagated through the CNN model $g_\theta(o_t)$ to construct the whole latent vectors $z_t$. The generative model architecture is presented in the appendix on github repository. $z_t$ is fed to the KF-RMLP to predict the labels (robot control actions). }
\label{Fig::din}
\vspace{-5mm}
\end{figure}

We exploit the recent advancements in Convolutional Neural Networks (CNN) in computer vision and Recurrent Neural Networks (RNN) to create an end-to-end solution for learning robot controller from a data-set of task demonstrations.
D-LfD automatically extracts features from the video of observations relevant to the task and maps the observations (video of the system including the robot interacting with a soft tissue) into robot's action. 
D-LfD includes four blocks, observation (OBS), feature extraction (FEX), augmentation (AUG), and control action generator (CAG), see Fig.~\ref{Fig::din}.
The raw visual inputs, i.e. video captured by a camera looking at a robot during execution of the task, are sampled to generate images $o_{i, t}$ at each control sampling time corresponding with $s_t$ and $a_t$. 
These images are fed to a CNN in OBS block to extract relevant features which are concatenated with robot's state and (possibly) with calibration parameter of the camera/s w.r.t. robot base frame, $\mathbf{T}^c$ in AUG block. 
Finally, the feature vector, which is the output from the AUG, is used by our RNN in CAG block to generate the control action at time $t$. 

OBS, FEX and AUG blocks form our \emph{feed-forward} model to generate the hidden state ${z}_t$ fed to the CAG block. 
We present different network architectures in the following for FEX and CAG blocks, e.g. Long Short Term Memory (LSTM), Gated Recurrent Unit (GRU), Simple Recurrent Neural Network (Simple RNN). Moreover, we propose a recurrent multi-layer perceptron trained by KF. 

\textbf{Feed-forward model:} We first use a a feed-forward network (CNN + dense layers) to learn a controller model for the robot from sample task executions.
This model extracts necessary features (latent feature vector) $z_t$ from image data by several convolutional and max-pooling layers concatenated with camera calibration parameters it feeds the latent feature vector into several dense layers to generate robot control input. 
Accordingly, the latent feature vector is estimated by applying the whole data sequence to the discriminative model. 
The learned model, hence, generates $\mathbf{u}_t$ (where $\mathbf{x}_{t+1}= \mathbf{x}_t+\mathbf{u}_t$) based on $\mathbf{x}_t$ and $o_t$. 
We have tested a series of state-of-the-art CNN architectures, including ResNet, VGG, and GoogleNet, to obtain the best architectures yielding the minimum error in predicting the control input in the test dataset where the corresponding results are presented in Table~\ref{table:cnn}. 
We also proposed a feed-forward model architecture, called F-CNN, which yields a performance better than other feed-forward models presented in Table~\ref{table:cnn}. 
Details of F-CNN is presented in the dataset repository. 
The latent feature vector of the best CNN is used for time series generation using the recurrent networks in the following sections.
\newline\textbf{Bootstrap Aggregation for Regression:}
Bootstrap Aggregation (BA) is usually used to improve the bias and variance of the model prediction on a test set. 
BA, also known as Bagging, is a meta-algorithm that reduces the variance of the statistical prediction of complex models such as deep neural networks. 
We utilise \emph{bagging} in the training to improve the variance of of the predicted values in test set error (reported in mean squared error). Moreover, bagging allows us to compute prediction confidence interval. 

We use bootstrapping (resampling with replacement) to generate $m$ new sets of data which are approximately independent and identically distributed ($iid$).

The variance of trajectories generated by ensemble model is $m$ times smaller than the variance of the trajectories generated by baseline model where $m$ is the number of ensemble cluster. 
In bootstrap aggregation we use re-smapling with replacement which is a method to approximate the true distribution $P$ by $\hat P$ to generate new sets of data from the training set. The \textit{resample} method from \textit{sklearn} library with a random \textit{random\_state} varying from 1 to 42 is used for re-sampling the data. After training models on each subset, we average the predictions to achieve the ensemble model which has (it is proved to have) lower variance.
We use bootstrapping to create 5 data sets and train 5 models on each one (weak learner). 
In precision tasks, e.g. needle insertion, reducing the prediction variance of the model enhances the reliability of the control system. 
%

\subsection{KF-Recurrent Multi-Layer Perceptron}
We use (Extended) Kalman filter (\textbf{KF}) which is a Gaussian state estimator (GSE) to train a recurrent multilayer perceptron (\textbf{RMLP}), called KF-RMLP. This model generates optimal control inputs, $\mathbf{u}_t$, from the latent state $z_t$ where the weights of RMLP form the states of the KF and has naturally a highly non-linear dynamics. 
This model also generates a confidence interval on the output signals. 

RMLP is a feedforward network with added time-delayed recurrent connections. 
We use the concept of Extended Kalman Filter to train the weights of the RMLP.

The network's behaviour can be described by the following equations:
\begin{equation}
W_{k+1} = W_{k} + \omega_k 
\vspace{-10pt}
\label{processequation}
\end{equation}
\begin{equation}
\mathbf{y_{k}} =  h_k (\mathbf{W_k}, \mathbf{u_k}, \mathbf{\sigma_{k-1}}) + v_{k} \label{observationequation}
\end{equation}
Eq.~\ref{processequation} is the process equation and shows that the state of the system is determined by a stationary process with network's weights values $W_{k}$ added up with process noise $\omega_k$. 
Equation \ref{observationequation}, known as observation equation, shows that the network’s target vector $y_k$ is a nonlinear function of the input vector $u_k$, the weight parameter $W_k$, and the recurrent
node activation $\sigma_k$; a random measurement
noise $v_k$ has been added up to this equation. The process and measurement noise are defined as zero-mean white noise with covariance given by  $E[\omega_k \omega_l^T] = Q_k$ and $E[v_k v_l^T] = R_k$ respectively.

The objective of the training problem for applying KF is defined as finding the global minimum loss function (mean-squared error) for the estimate of the state $\hat{W}_k$, using all observed data to that point.
The KF solution pseudo-code to the training problem is illustrated by Algorithm \ref{KFAlgorithm}.

For initialization, network weights are set as small random values drawn from a zero-mean normal distribution. 
The estimation of the state (i.e., weights) of the system at step $k$ is presented by the vector $\hat{W}_{k}$.
This estimation is a function of Kalman gain matrix $K_k$ and the error vector $\xi_k = y_k - \hat{y}_k$, where the $y_k$ is our labelled target vector and $\hat{y}_k$ is the estimation of the target vector, which is the network's output, at $k$th step.

\begin{algorithm}[t!]
\caption{Extended Kalman filter-Recurrent Multi Layer Perceptron}\label{tab:kf}
\begin{algorithmic}[1]
\Procedure{KF-RMLP}{}
\State  \textit{Initialisation} for $P_k, R_k,$ and $Q_k$ at {\small $k=0 $, 
}
\State For $k$ = 1, 2, ..., \emph{T}:
\State {\small Update error vector by forward propagation: \hfill \newline. \hfill $ \mathbf{\xi}_k = \mathbf{y}_k -  \mathbf{\hat{y}}_k $}
\State {\small Update derivative matrix $\mathbf{H}_k$ by back propagation}
\State {\small Update global scaling matrix: \hfill \newline. \hfill $ \mathbf{A}_k =[ \mathbf{R}_k +  \mathbf{H}_k^T  \mathbf{P}_k  \mathbf{H}_k]^{-1} $}
\State {\small Update Kalman gain matrix: \hfill \newline. \hfill $\mathbf{K}_k = \mathbf{P}_k  \mathbf{H}_k  \mathbf{A}_k$}
\State {\small Update Kalman state (weights of NN): \hfill \newline. \hfill $\hat{\mathbf{w}}_{k+1} =\hat{\mathbf{w}}_{k+1}  +\mathbf{K}_k  \boldsymbol{\xi}_k $}
\State {\small Update approximate covariance matrix: \hfill \newline. \hfill $\mathbf{P}_{k+1} = [I - \mathbf{K}_k \mathbf{H}_{k}] \mathbf{P}_{k}$ }
\EndProcedure
\end{algorithmic}
\label{KFAlgorithm}
\end{algorithm}

The Kalman gain matrix is a function of: 1. approximate error covariance matrix $P_k$ which is initialized by $\epsilon^{-1}I$, where $\epsilon=0.001$ and $0.1$ for linear and sigmoid activation respectively; 2- matrix of derivatives of the network’s outputs w.r.t all trainable weight parameters $H_k$ and 3- scaling matrix $A_k$.
The scaling matrix $A_k$ is constructed with a
function of the measurement noise covariance matrix $R_k$ which is set as $\eta^{-1}I$ and $\eta$ is the learning rate, the
matrices $H_k$ and $P_k$. Finally, the approximate error covariance matrix $P_k$
is updated recursively with the weight vector estimate; this matrix includes
second derivative information about the training problem, and is augmented
by the process noise covariance matrix $Q_k$ which is equal to $q_{k}I$ and $0<q_{k}<0.1$. The goal of this algorithm is to find weights values, which will minimize the value of the loss function $\Sigma  \xi_k^T \xi_k$.
The process and observation noise covariance matrices, $R_k$ and $Q_k$, should be specified for all training
instances. Similarly, the approximate error covariance matrix $P_k$ must be
initialised at the beginning of training. The training procedure can be described by the following steps: 1. Forward propagation and error vector calculation to have $\hat{y}_k$ and $\xi_k$; 2. Backpropagation to calculate $H_k$ which is output error's derivative with respect to all trainable weights. Backpropagation includes the information about a time history of recurrent node activations for RMLPs; 3. Kalman gain calculation by having $H_k$, approximate error covariance matrix $P_k$ and the measurement covariance noise matrix $R_k$; 4. Network wights update by having $K_K$, $\xi_k$ and the current values of the weights; 5. Updating approximate error covariance matrix.

\section{Experimental Evaluation}
\begin{figure}[tp!]
\centering
\includegraphics[width=0.47\textwidth,height=30mm]{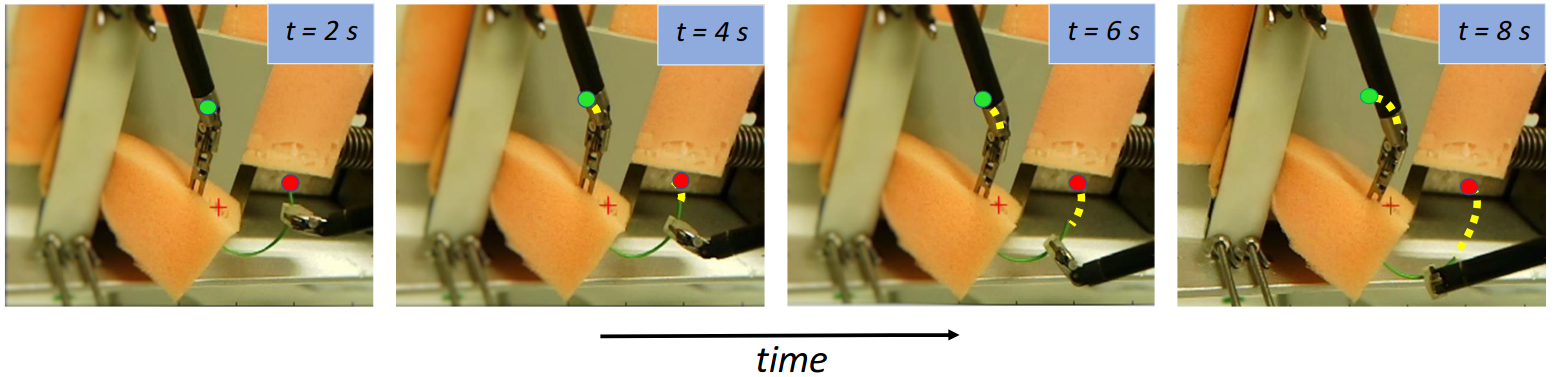}
\vspace{-5pt}
\caption{Sample frames of the training sequences. The red and yellow dots are shown as fixed points in image frame to depict the motion of robot's left and right arms. The yellow dot curve shows their approximate motion path. Our goal is to track the red cross sign and predict the coefficients of robot controller at the same time.}
\setlength{\intextsep}{-10pt}
\label{fig::videtimelapse}
\vspace{-5pt}
\end{figure}

To validate our proposed D-LfD control architecture, we generated a data-set of needle insertion in a soft, deformable object (see Fig.~\ref{fig::videtimelapse}). 

\textbf{Robot Surgery data-set}--Our data-set contains 60 successful needle insertion performed by two arms of a da Vinci Research Kit robot. 
Right arm holds the needle and pushes it through the soft (PVA sponge) tissue while the left arm manipulates the objects to ensure the needles passes through the desired exit point, which is crossed with red colour in the video feed (see Fig~\ref{fig::videtimelapse}). 
Each experiment is recorded by 3 pairs of stereo-cameras-- each provides 2 views--giving us 6 video recordings of each task. 
Videos are synced with the corresponding robot kinematic data. Fig~\ref{fig::videtimelapse} shows a sample sequence of four images captured by a camera. 
For preprocessing of videos, we cropped the videos where image frames are generated by the sampling time similar to the ones of kinematic data.  
Average time-to-completion is 7 [s] where the total frames of videos are generated by the constant sampling frequency. The total samples of each experiment vary across different experiments.  
We have, in general, 90,000 sample images for the 60 experiment videos. 
The data-set is split into 80\% and 20\% for training and testing respectively and 30\% of training data are used for cross-validation. 
We have used Tensorflow framework for training with Adam optimiser for gradient update and Lasso regulariser with $\lambda = 0.1$. Our proposed recurrent model (KF-RMLP) is trained via (Extended) Kalman Filter algorithm instead. 

The control input which our D-LfD network outputs consist of 7 components for the left hand which manipulates the soft tissue, $\mathbf{u}_t = \{q_{t+1}, p_{t+1}\}$ where $q_{t+1}$ and  $p_{t+1}$ are orientation expressed by quaternion and position of the end-effector at $t+1$. 
The right hand follows a predefined trajectory where the exit point is randomly selected in a region for each experiments. Hence, $e = \|\hat{p} -p\| +d(\hat{q}, q)$ which express Euclidean and quaternion distances of predicted and ground truth control input to the robot.
\vspace{-0mm}

\paragraph{Feed-forward model}--We trained different SOTA feed-forward networks, e.g. LeNet, AlexNet, VGG19, ResNet, and GoogleNet, in combination with Dense layers to predict the control inputs. As Table~\ref{table:cnn} shows our feed-forward model outperforms others in terms of prediction accuracy and error values. 
%
\begin{table}[tb!]
\caption{Evaluation of tested CNN architectures. F-CNN yields best performance.}
\centering
\vspace{-3mm}
 \begin{threeparttable}
 \vspace*{-\baselineskip}
\begin{tabular}{ c c c c c c}
 \hline
 \textbf{Model}& MAE \tnote{$\dagger$}&  AE \tnote{$\ddagger$}& Loss& E $>$ 0.01 \tnote{$\ddagger\ddagger$}\\
 \hline 
 LeNet   &  0.5982    & 0.3845& 0.3347& 98.57\\

 AlexNet   &   0.3954    &  0.2651&  0.1478&  86.20\\


 VGG19   & 0.1334    &  0.0187&  0.0201& 26.54 \\
 
 ResNet18   & 0.0984    &  0.0187&  0.0158& 27.43 \\
 
 GoogleNet   & 0.1458    &  0.0156&  0.0130& 27.65 \\
 
 \textbf{F-CNN}   & \textbf{0.0947}  &  \textbf{0.0136}&  \textbf{0.0137}&  \textbf{27.03} \\
 \hline
 \end{tabular}
 \begin{tablenotes}
     \item[{$\dagger$}] Maximum Absolute Error \item[{$\ddagger$}] Average Error
     \item[{$\ddagger\ddagger$}] No. of Error elements $>$ 0.01 (\%)
   \end{tablenotes}
 \end{threeparttable}
\label{table:cnn}
 \vspace*{-5mm}
\end{table}

\begin{figure}[h]
\subfigure[\small CNN]{\includegraphics[width=.45\columnwidth]{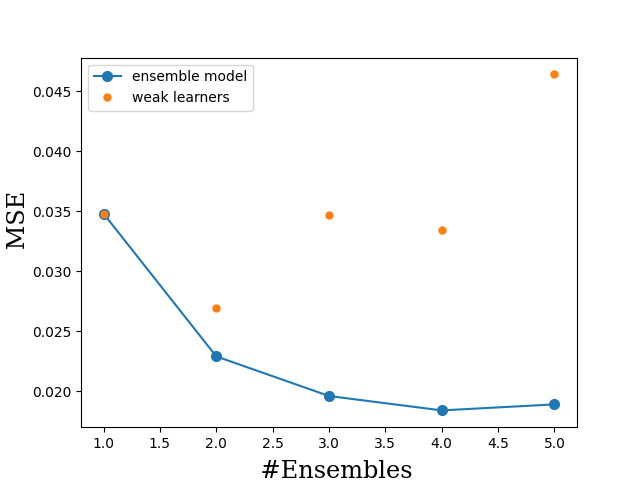}
  \label{fig:sfig19}
}
\hspace{-4.5mm}
\subfigure[\small RNN]{  \label{fig:sfig20} \includegraphics[width=.45\columnwidth]{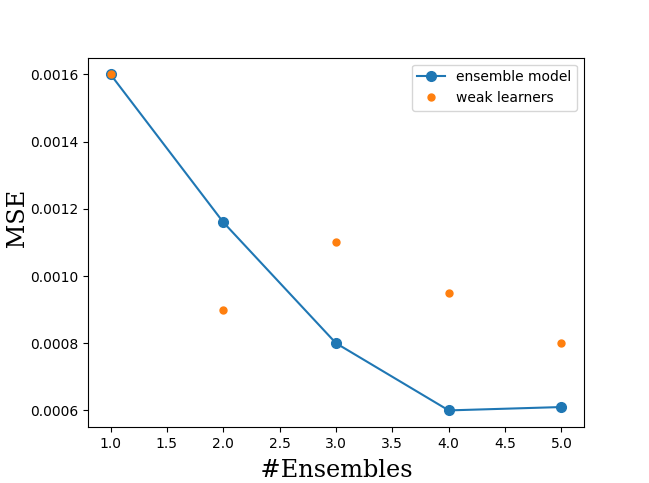}
  }
\caption{test set MSE of the weak learners and ensemble models}
\label{fig:MSEensemble}
\end{figure}

We have selected the best model which is the customised CNN as feature extractor in deep time series prediction with recurrent networks and also for base ensemble model in bagging. Moreover, to improve the accuracy of the prediction we tested novel discriminative architectures by tuning the hyper-parameters and the network with the best performance are selected.
\paragraph{Feed-forward model with bootstrap aggregation}Bagging is a technique for model uncertainty evaluation and reduce variance of complex models. Bagging improved prediction performance (see Table \ref{table:kfrmlp}) and MSE (figure \ref{fig:sfig19}) relative to a base CNN model trained on all of the dataset.
By bootstrapping 5 new dataset were generated and the case of 4 ensembles yielded best performance. In theory, the error should converge to zero as the number of ensembles increases, but creating more data sets by bootstrapping from a limited amount of data leads to stronger correlation between the ensembles which contradicts the basic assumption in bagging (weak learners independence) and increases the MSE. 
\vspace{-0mm}

\paragraph{D-LfD results}
Table 3 present the error metrics including the percentage of the samples with error more than 1\% (E$>$0.01) of different tested recurrent networks. 
The recurrent networks significantly improved the performance, in terms of all error metrics, compared to feed-forward model, their prediction errors are very close.
Interestingly our proposed network, namely KF-RMLP, outperforms all the other networks with significantly better error values.   
An interesting observation is that E$>$1\% error for KF-RMLP is quite smaller than others which suggest our model can be trusted more.

\begin{table}[tb!]
\vspace{0pt}
\centering
 \begin{threeparttable}[b]
\caption{Evaluation of KF-RMLP and comparison with LSTM, GRU and Ensemble models. The loss function is MSE and our approach outperforms in all of these mentioned terms.}

\begin{tabular}{ c c c c c c c  }
 \hline
 \textbf{Model}&  MAE&   AE&  Loss&  E $>$ 0.01 \\
 \hline 
 \hline

 feed-forward   &    0.094&   0.0136&   0.0137&   27.03\\ 
 
 LSTM   &       0.031&   0.0065&   0.0054&   7.18\\

 GRU   &       0.024&   0.0061&   0.0059&   9.67\\

 RNN   &      0.021&    0.0066&   0.0123&   15.34\\
 
 CNN-Ensemble&  0.080&  0.0083&  0.0109&   22.71 \\
 
 GRU-Ensemble & 0.019&  0.0048&  0.0049&   5.12 \\

 \textbf{KF-RMLP}   &      \textbf{0.0014}&   \textbf{0.0013}&   \textbf{0.0018}&   \textbf{3.43}\\
 \hline
\end{tabular}
\label{table:kfrmlp}
 \end{threeparttable}
  \vspace*{-5mm}
\end{table}


The results in Table~\ref{table:kfrmlp} illustrate that our approach outperforms both the standard feed-forward model and LSTM as well as GRU estimators. In the feed-forward model, the tracking error (controller prediction error) is large due to the lack of appropriate state estimator. On the other hand, our approach's performance not only depends on the observations, but it also learns to find a compromise between the observations and the dynamic model due to its recurrent structure.
The optimisation methods in GRU and simple RNN cells depend on instantaneous estimations of the gradient: the derivatives of the error function w.r.t the weights to be adjusted only based on the distance between the current output and the corresponding target, using no history information for weight updating.

Although the LSTM model is the most general and state of the art, it does not perform better than our method, since it does not incorporate prior knowledge about the structure of the state estimation problem. In addition, the proposed optimisation method for LSTM is likewise the optimisation method of GRU and Simple RNN, so no history information is used for weight updating. The (Extended) Kalman Filter overcomes this limitation. It considers training as an optimal filtering problem, recursively and efficiently computing a solution to the least-squares problem of finding the curve of best fit for a given set of data in terms of minimising the average distance between data and curve. At any given time step, all the information supplied to the network up until now is used, including all derivatives computed since the first iteration of the learning process. However, computation is done such that only the results from the previous step need to be stored.

\begin{figure}[tb!]
\subfigure{\includegraphics[width=.47\columnwidth]{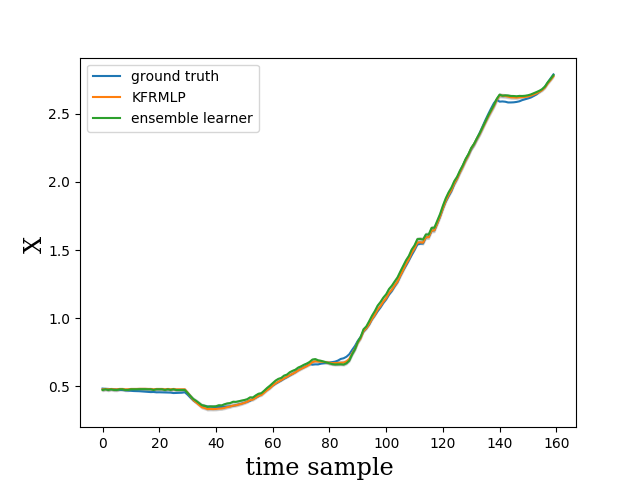}}
\hspace{-4.5mm}
\subfigure{ \includegraphics[width=.47\columnwidth]{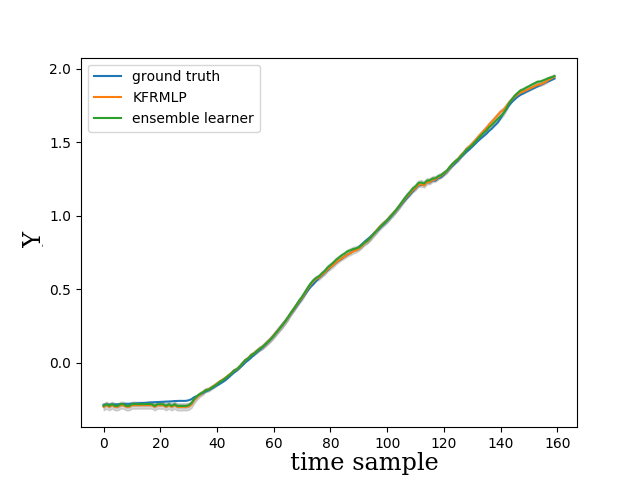}
}
\vspace{-4mm}
\hspace{-4.5mm}\\
\subfigure{\includegraphics[width=.47\columnwidth]{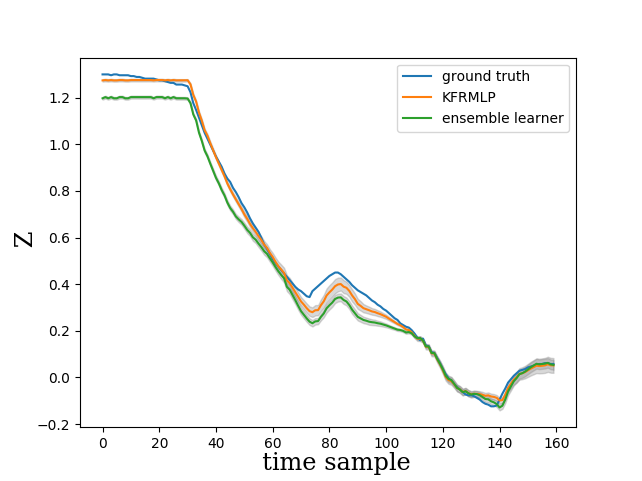}
}
\hspace{-4.5mm}
\subfigure{\includegraphics[width=.47\columnwidth]{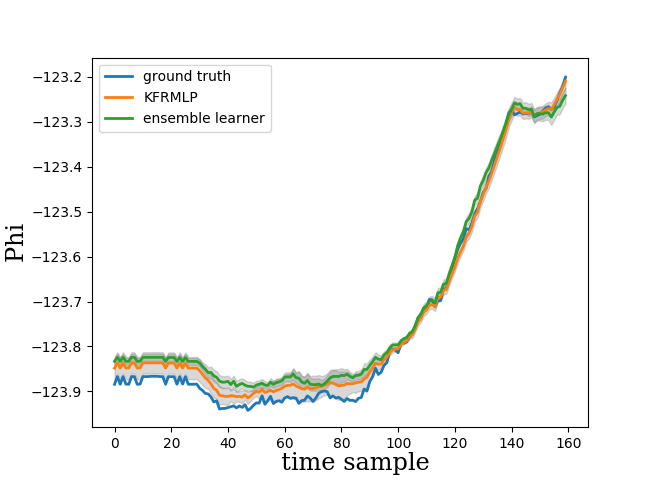}
}
\hspace{-4.5mm}\vspace{-4mm}
\\
\subfigure{\includegraphics[width=.47\columnwidth]{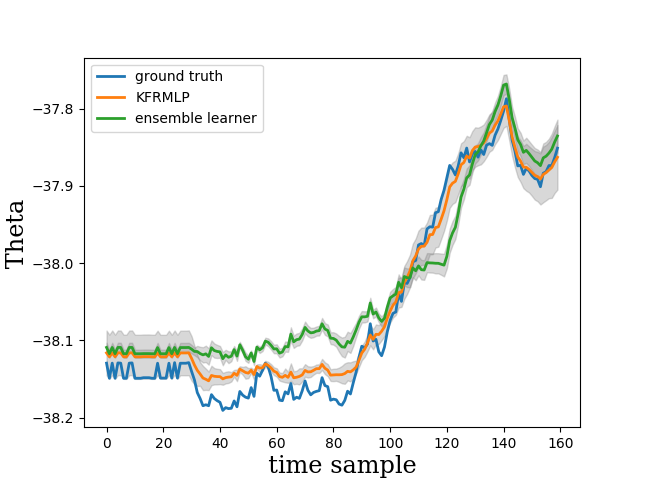}
} 
\hspace{-4.5mm}
\subfigure{\includegraphics[width=.47\columnwidth]{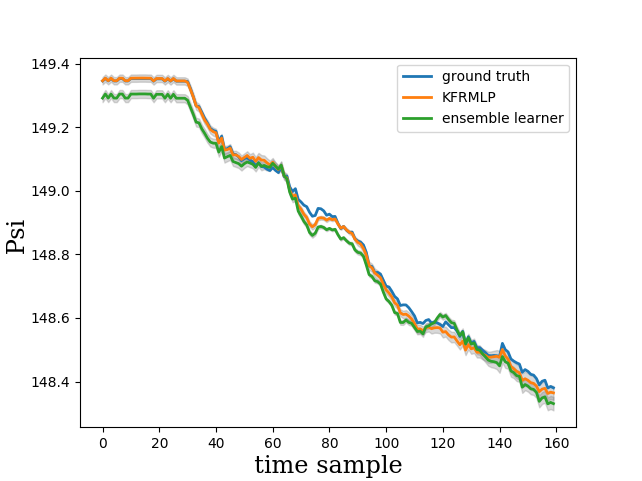}}
\hspace{-4.5mm}
\vspace{-4mm}
\caption{KF-RMLP and RNN output for a sample task trajectory.}
\label{fig:KFRMLPandEnsemble}
\end{figure}

\paragraph{Recurrent network with Bagging Ensembles} We have also used bagging on GRU model with five ensembles to compare the results with KF-RLMP. The results in Table \ref{table:kfrmlp} indicate the ensemble GRU model has better performance relative to the base GRU trained on all of the training data. 
It is clear that additional to significant improvement in prediction accuracy relative to CNN models (see also figure \ref{fig:KFRMLPandEnsemble}), the confidence interval of the recurrent ensemble model is very narrow which indicate high confidence and low variance for testing on unseen test data.

Figure \ref{fig:KFRMLPandEnsemble} illustrates the relative performance of the ensemble GRU model and the proposed KF-RMLP. As the figure shows, KF-RMLP has closer trajectory to the ground truth and smaller tracking error values. Further to higher prediction accuracy the confidence interval of the KF-RMLP is relatively narrower w.r.t GRU-ensemble model which denotes the higher reliability of the model.

 \section{Conclusion}
In this paper, we presented a novel dataset useful for Deep-Learning from Demonstrations (D-LfD) that allows a robot to learn control policy to perform a complex task only from observations. This is quite important for many robotic tasks which are fully teleoperated. For example, despite the significant amount of data available for robotic surgery, this task is still fully teleoperated. 
We also provide a few baseline implementations of D-LfD which outperforms state-of-the-arts. 
We tested different recurrent networks including LSTM, GRU and RNN. Moreover, we proposed KF-RMLP which is recurrent MLP trained by an extended Kalman filter.  Our experimental results show that our proposed KF-RMLP outperforms other state-of-the-art recurrent networks and yield a very low value of errors. 
Furthermore, the number of instances our KF-RMLP network generates with errors more than 1\% are significantly lower than other networks. 

Although the proposed D-LfD needs further development for real-robot deployments, they can already be used for training novices via a haptic-guided shared control where the haptic guidance can be generated based on the control input generated by D-LfD network and the stiffness of the guidance can be modulated using the confidence interval generated by our network. 
Moreover, the dataset helps us study several related problems. 
For instance, the dataset allows us to study how a controller learned based on Camera view 1 can be generalised to camera view 2. This is very important as the dataset collected during different operations are captured by different camera views.  
Future works include test D-LfD on real robot experiments, generalising the learned controller to different camera view and generalising to tissue manipulation with different mechanical properties. 
\vspace{-5pt}


\bibliographystyle{ieeeconf}
\bibliography{reference}

\end{document}